\documentclass[11pt,a4paper]{article}
\usepackage{authblk}
\usepackage{acl2015}
\usepackage{times}
\usepackage{url}
\usepackage{graphicx}
\usepackage{pifont}
\usepackage{csquotes}
\graphicspath{ {images/} }

\title{Planned Event Forecasting using Future Mentions and Related Entity Extraction in News Articles}

\author[ ]{\textbf{Neelesh K Shukla}}
\author[ ]{\textbf{Pranay Sanghvi}}
\affil[ ]{Department of Computer Science and Engineering}
\affil[ ]{Indian Institute of Technology, Guwahati}
\affil[ ]{\{neelesh.shukla,s.pranay\}@iitg.ernet.in}

\date{}

\begin{document} 
\maketitle

\begin{abstract}
  In democracies like India, people are free to express their views and demands. Sometimes which causes the situation of civil unrest like protest, rallies, marches etc. These events may be disruptive in nature and most of the times held without taking prior permission from competent authority. Forecasting these events will help administrative officials to take necessary action. Usually, protests are announced well in advance to have large participation. Therefore by going through these kind of announcements in news articles, such planned events could be forecasted beforehand. We developed such a system in this paper to forecast social unrest event by using topic modelling and word2vec to filter relevant news articles,  Name Entity Recognition (NER) method for identifying entities like people, organization, location and date. Time normalization is used to convert future date mentions in standard format. In this paper have developed a geographically independent generalized model to identify keys for filtering civil unrest events. There could be many mentions of entities and very few may be actually involved in the event. This paper calls such entities, 'Related Entities' and proposes a method to extract such entities referred as 'Related Entity Extraction'.
\end{abstract}

\section{Introduction}

In democratic countries like India, civil unrest is a common happening. These events could be disruptive and hence a beforehand forecasting could help officials to maintain law and order. There can be two flavours: Planned and Unplanned. This paper has focused on planned event forecasting. As per our hypothesis, more people need to be involved for protest and for such large mobilization, protest should be announced in advance. We can identify such events and extract the relevant information for forecasting.\\
\-\ \-\ To build a protest forecasting system, our first aim was to find all protest related information. There are multiple channels of communication for such information like news websites, blogs, social networking sites etc. We have selected news feeds to look for people, organization, date and location associated with the protest. There could be many entities in a news article, so the major challenge is to find entities which are actually involved in the event. Very less work has been done in this area. \\
\-\ \-\ In order to build our data set, we collected news articles from various leading electronic media of India. Then we applied article filtering using phrase learning combined with topic modelling approach. Identifying entities which are actually related to the event is a major problem in such kind of system still there is no good solution for it. This paper has focused on providing the solution for it. Previous works like Muthiah et al 2015 and others haven't addressed this issue and focused only on time and location. We have identified entities using Named Entity Recognition and related entities using relation extraction combined with our lexicon and window based model that we are going to propose.\\
\-\ \-\ The rest of paper is organized as follows. Next section discusses related works. Section 3 describes our approach of dataset building, relevant article identification, related entity extraction and time normalization. Section 4 includes experiments and results and Section 5 concludes this paper.

\section{Related Work}

In recent times, Event forecasting has been attracting researchers around the globe. Event forecasting has been studied for multiple domains and civil unrest is one of them. Mostly they used data which is spatio-temporal in nature and various information extraction techniques to retrieve the required information. Event forecasting problem can be divided into various sub-problems such as event detection, information extraction, named entity recognition, relation extraction and time expression normalization etc. Event detection has become one of the emerging areas of study in the literature. Allan 2002, Yang et al 1998 and Gabrilovich et al 2004 have used document clustering techniques, where as Banko et al 2007 and Chambers et al 2011 focused on extracting patterns to extract information from the text. Another heavily studied topic is temporal information extraction. TempEval challenge Verhagen 2007 led to significant development of algorithms for temporal NLP. For instance, Pustejovsky et al 1991 described a specification language that can be used for evaluating  temporal expressions in natural language. A system named EMBERS as described in Ramakrishnan et al 2015, forecasts civil unrest events using open source indicators and focused on shallow mining of broad set of data sources where as we focused on planned events. \\
\-\ \-\ We found two publications which align closely to our work, Compton et al 2013 and ~\cite{} Xu et al 2014 as their emphasis is on protest forecasting like ours. Both of them have used different  data source and different methodology to forecast the protests. Compton et al 2013 finds the future date by searching for keyword that mentions date in absolute terms i.e. finding name of month and a number less than 31 which represents date. This approach won't be able to extract phrases used to reference future dates like \enquote{Day after Tomorrow}, \enquote{Next Wednesday} etc. In Xu et al 2014, The same group of authors have worked using Tumblr feed on  much smaller collection of keywords but again that work is limited to use of absolute time identifiers.\\
\-\ \-\ Similar work has been done primarily on Latin America region by Muthiah et al 2015.  They forecasted the location and date of a planned protest event but they failed to give information about people or organization involved in the event. We have focused on social unrest events like protest, rally etc and used content based approaches for our methods. We have focused on spatio-temporal expression present in news text to locate and forecast event. Unlike existing works, we have developed methods for identifying Related Entities which means entities involved in the event.

\begin{figure*}[h]
\includegraphics[width=\textwidth]{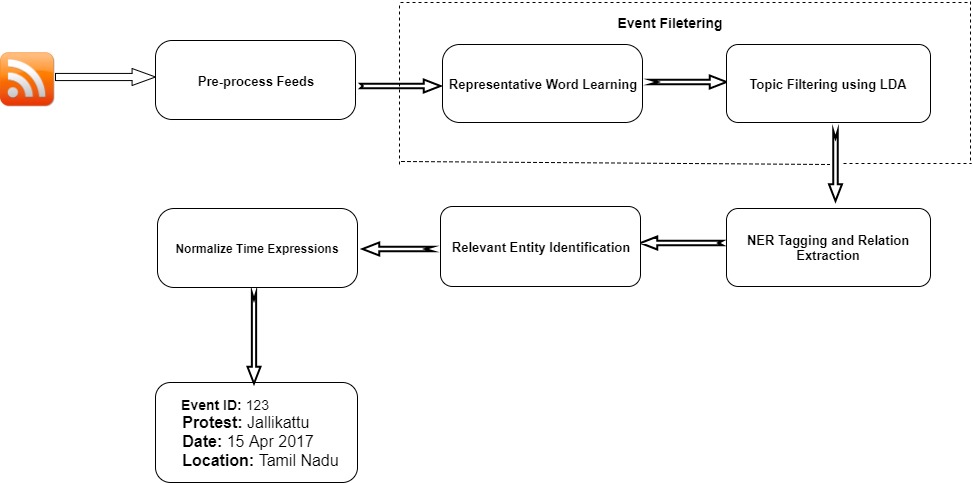}
\caption{Overall flow of our system}
\end{figure*}

\section{Approach}
In general, our approach was to get the news articles from RSS feeds, pre-process them, identify the articles related to social unrest and extract relevant information like person, organization, time and location to forecast event. Fig 1. shows our overall approach.
\subsection{Building Corpus}
First step to build the dataset was to parse the news feeds from multiple news websites. We chose four news websites viz Hindustan Times, Times of India, The Hindu and Indian Express as they were publishing significant number of articles per day.

\subsubsection*{Fetching the News Articles and Pre-processing}
We have parsed the news feeds from 16\textsuperscript{th} January 2017 to 25\textsuperscript{th} March, 2017 and extracted content, publication date and title of the article. The content of article was passed through shallow pre-processing as tokenization, lemmatization (normalizing words for inflection), removal of punctuation and stop words. 

\subsection{Relevant Document Identification}
The next task was to identify the articles relevant to the event in which we are interested. We have done this in two steps. \\
\-\ \-\ i) Learn the keys and filter the article based on presence and absence of these keys. 
\-\ \-\ ii) Refine the filtered articles using topic modelling.

\subsubsection*{Representative Word Learning}
In general there are multiple words or phrases which can be used to address domain specific events for example instead of using word protest, different articles can use its synonyms or words used in similar context as protest. In word learning, we need to list out all such words. For this initially we identified seed words for social unrest domain based on our domain knowledge which are PROTEST, DEMONSTRATION. Similar method has been used by Muthiah et al, 2015. \\
\-\ \-\ There can be many other words which may be used in the similar context. There are two ways to learn all such words. One way is to find all the synonyms of protest and demonstration. This can be done by using WordNet\footnote{WordNet, Princeton University; https://wordnet.princeton.edu/}. This approach has two problems. One, the synonyms may not be relevant like presentation, test, seminar are synonyms of demonstration but are not related to events of social unrest. Another issue is that synonyms may miss the domain specific terms or the terms used in local languages like Dharna, Bandh etc.

\-\ \-\ Another way is to find all the words which share the same context as seed keywords. We used word2vec model. word2vec model needs to generate word embedding for each word. We used our collected news articles as corpus to generate word embeddings. We took `protest' and `demonstration' as target words and generated words having similar context. We identified the relevant documents which contained these word phrases. As shown in results section, we were able to get the words in local language like `bandh', `dharna' and other words like `agitation', `march' which are similar to our seed words and relevant. This way, we got generalized keywords.

\-\ \-\ We assumed that presence of one of these learned words, makes document relevant. With that assumption, filtering on presence or absence of a word in article may pick the articles which are relevant to us as well as few documents which are not relevant to us. For example, a word `strike' will fetch the articles related to protest as well as surgical strike. So the filtered documents should be refined further. Topic modelling was used to achieve the same.

\subsubsection*{Document Filtering using Topic Modelling}
LDA (Blei, Andrew et al, 2003) model is used to identify hidden themes (topic) in document. Each document can be thought of as mixture of topics and each topic can be considered as mixture of words. \\

\begin{figure}[h]
\includegraphics[width=\linewidth]{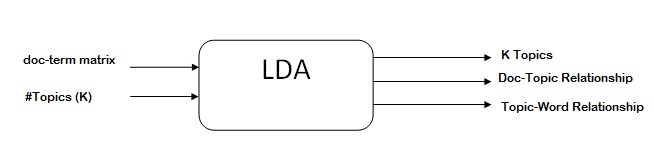}
\end{figure}

\-\ \-\ We used vector space model (VSM) representation of doc, which is very frequently used in solving many kind of Information Retrieval problems. In vector space model a document can be represented as n-dimensional vector where n is vocabulary size. d(w\textsubscript{1}, w\textsubscript{2},...., w\textsubscript{n}) Each entry d\textsubscript{ij} is weighed term frequency of a word w\textsubscript{j} in document d\textsubscript{i}.

To use LDA for our purpose of document filtering, we followed below procedure: \\
\-\ \-\ Initially for all the key phrase filtered document, we removed less frequent words since clearly these words would not be having much contribution in document. We built term-document matrix and calculated TF-IDF to give weights to terms according to their importance in document. We passed this matrix to LDA API of  gensim\footnote{Gensim, https://radimrehurek.com/gensim/} and generated  topic-word mixture. We exploited topic-doc relationship to filter relevant documents to the chosen topic. More details on implementation are explained in Experiment and results section.

\subsection{Related Entity Extraction}
To forecast an event, attributes of the event need to be extracted. Our goal was to get the following attributes.\\
{[Date, Location, Person, Organization]}\\
\-\ \-\ Named Entity Recognition (NER) was used to extract all the entities with tags Date, Time, Location, Person and Organization. We used Stanford Named Entity Recognizer\footnotemark (Finkel et al, 2005) which is also known as CRFClassifier, implements linear chain Conditional Random Field (CRF) sequence models. It is 7-class {Location, Organization, Date, Money, Person, Percent, Time} classifier for  English language.\\
\footnotetext{Stanford Named Entity Recognizer, https://nlp.stanford.edu/software/CRF-NER.shtml}

\-\ \-\ An article contains many entities which may not be actually involved in protest. For example, a line like, `Narendra Modi\textless PERSON\textgreater \-\  called a meeting to discuss issue raised in protest' has person tag, but Narendra Modi is not actually involved in protest. We wish to get those entities which are actually involved or closely related to the event. In this paper, we refer these entities as `Related Entities'.\\
\-\ \-\ To get these related entities we are going to introduce a `Related Entity Extraction' model. Our model first extracts following relation pattern.

\textless Part1 containing Entity1, Verb Phrase, Part 2 containing Entity2\textgreater

\-\ \-\ We used Stanford Relation Extractor to build these kind of triplets which implements algorithm described in paper (Surdeanu et al, 2011) and generated relations like:\\
(jpp\textless organization\textgreater, call on, february\textless date\textgreater \-\ 16\textless date\textgreater) \\
(jpp\textless organization\textgreater, call for, Jharkhand \textless location\textgreater bandh)

\-\ \-\ Once these relations are built, related entities need to be extracted. For this purpose, we designed a 2-step method that we are going to explain further. By analyzing the articles we found that related entities occur near by. We can exploit this property of spatial locality to extract Related Entities. For example, see the above two nearby relation triplets which have date and location together. Apart from these relations, the title of a news article has high chance of having related entities. We utilized these unique features to build our method called, `Window Based Related Entity Extraction' to get related entities. Idea is that if we can locate one related entity, we can locate other related entities nearby. We applied lexicon based  approach and looked for certain patterns in these relation triplets like `call on' \textless DATE\textgreater etc to locate first related entity. We used dependency parser idea of Muthaiah et al, 2015 to learn these patterns. Following methods is used to identify related entities at document level.\\

\textbf{Related Entity Extraction Method}
\begin{enumerate}
  \item Build and learn patterns for the phrase which will be used to identify relations having related entities.
  \item Create a separate list of each class of entity.
  \item Start processing relation triplets sequentially.
  \item Pick the relation and check for phrase. If relation follows the phrase pick that relation and relations in its window.
  \item Locate entities, if no entity present in the relation, mark that relation processed and go for next relation.
  \item If relevant entity present extract person, date, location and organization.
  \item Check if the entity is already present in its respective list and if so, ignore entity, otherwise add them in their respective list and mark relation processed.
  \item Continue above steps till there is any unprocessed relation
  \item Mostly titles of news articles have most relevant information, check tagged title of article for the patterns defined in first step and extract information. 
\end{enumerate}
We ignored co-reference and entity disambiguation issues.

\section{Experiment and Results}
In this section, we are going to explain our dataset, different parameters that have been used at each stage, evaluation methods and results.

\subsection{Dataset}
Real world news articles were crawled from Times Of India, Hindustan Times and Indian Express starting from 16\textsuperscript{th} January 2017 and till 25\textsuperscript{th} March, 2017 and 26219 articles were downloaded.

\subsection{Relevant Document Identification}
We started with application of LDA on our main corpus. Since number of protest related articles were very less, we were not getting good results. LDA was returning all the political articles which were noise for us.

\-\ \-\ To resolve this issue, we focused on improving the density of protest articles and started working on phrase based filtering. We first tried synonyms but didn't get good results as we described earlier also in section 3.2. We used Word2Vec model to get similar words to PROTEST and DEMONSTRATION with cut-off value 0.68.

\begin{figure}[h]
\includegraphics[width=\linewidth]{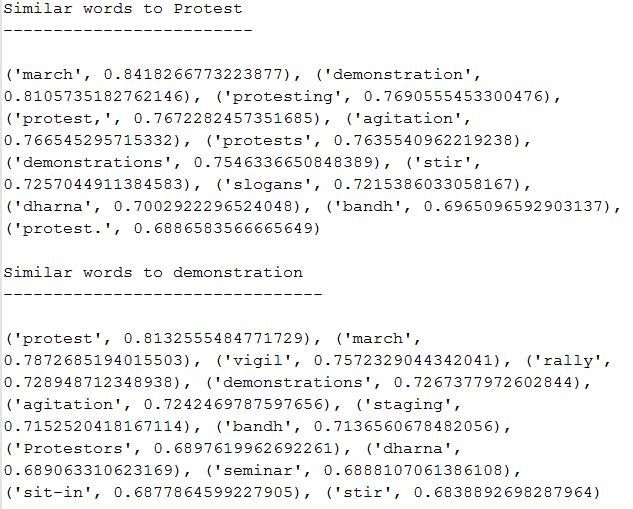}
\end{figure}

\-\ \-\ Simply presence and absence of a keyword doesn't guarantee relevance. For example: Strike could relate to surgical strike or Russian strike on Ukraine which is not relevant to us as we are interested in the events of social unrest. We applied LDA on this filtered corpus to get more relevant articles having details of social unrest event. We used following parameters for LDA: symmetric alpha: 0.0067, symmetric eta: 6.89, Topics: 50, 100, 150, Passes: 20

\-\ \-\ We evaluated our relevant document identification model on precision, recall and f measure. To do so, 2500 articles were sampled from the main corpus and tagged with relevance true (1) and false(0). Since relevance of article is subjective, to have reliable tagging, we given these articles for tagging to an independent team. Final relevance was calculated by taking average by setting min threshold .5 for giving relevance as true. 
We achieved 85\% precision, 69.5\% recall and F-measure 76.6\%. 

\subsection{Related Entity Extraction}
We evaluated our related entity extraction algorithm in terms of relevant relations triplets. Relation triplets were tagged as relevant if they had related entities.  Tagging was again done by an independent team as discussed in previous part. We evaluated our model on precision, recall and accuracy measures.

We achieved 64.3\% precision, 63\% recall and accuracy 87\%.

\section{Conclusion and Future Work}
This paper presented a system which can forecast a planned event of social unrest by looking into content and identifying future mentions. In this paper, we majorly focused on relevant article identification and related entity extraction and ignored other issues like duplicate event removal, entity co-referencing and entity disambiguation. These can be incorporated later for further improvement.

As a future work, apart from giving entities involved in the event, purpose of that event could also be forecasted by using text summarization techniques. Since the social unrest event could be disruptive in nature, one can apply sentiment analysis techniques to define the intensity of the event, which can help authorities to plan according to that.

\end{document}